\PassOptionsToPackage{verbose=silent}{microtype}
\documentclass[]{fairmeta}
\usepackage{silence}
\usepackage[utf8]{inputenc} % allow utf-8 input
\usepackage[T1]{fontenc}    % use 8-bit T1 fonts
\usepackage[english]{babel}% or your language
\makeatletter
\renewcommand{\title}[1]{%
  \gdef\titlelist{{\fontsize{19}{23}\selectfont\bfseries #1}}%
}\makeatother

\usepackage{xspace}
\usepackage{wrapfig}
\usepackage{algorithm}        % the floating environment
\usepackage[noend]{algpseudocode}  % pseudocode macros (via algorithmicx)
\algrenewcommand\algorithmicrequire{\textbf{Inputs:}}
\algrenewcommand\algorithmicensure{\textbf{Outputs:}}

\usepackage{amssymb}
\usepackage{pifont}
\usepackage[most]{tcolorbox}

\usepackage{amsmath,amsfonts,bm}

\def\eqref#1{equation~\ref{#1}}
\def\1{\bm{1}}

\DeclareMathAlphabet{\mathsfit}{\encodingdefault}{\sfdefault}{m}{sl}
\SetMathAlphabet{\mathsfit}{bold}{\encodingdefault}{\sfdefault}{bx}{n}

\usepackage{hyperref}       % hyperlinks
\usepackage{url}

\usepackage{booktabs}       % professional-quality tables
\usepackage{amsfonts}       % blackboard math symbols
\usepackage{nicefrac}       % compact symbols for 1/2, etc.
\usepackage{microtype}      % microtypography
\usepackage{xcolor}         % colors
\usepackage{graphicx}
\usepackage{float}
\usepackage{subcaption} % use instead of the deprecated 'subfigure'

\usepackage{amsmath,mathtools}
\usepackage{amsthm}

\usepackage{array}
\usepackage{multirow}
\usepackage{tabularx}
\usepackage{threeparttable}
\usepackage{colortbl}
\usepackage{siunitx}

\usepackage{newfloat}
\usepackage{listings}
\DeclareCaptionStyle{ruled}{labelfont=normalfont,labelsep=colon,strut=off}
\floatstyle{ruled}
\newfloat{listing}{tb}{lst}{}
\floatname{listing}{Listing}
\usepackage{enumitem}
\usepackage{cancel}
\usepackage{soul} % for \hl etc.

\usepackage{etoc}

\theoremstyle{definition}

\renewcommand{\epsilon}{\varepsilon}

\theoremstyle{plain}
\theoremstyle{remark}

\usepackage{bm} % for \bm
\newtcolorbox{promptenv}[2][]{%
  enhanced,
  breakable,
  colback=white,
  colframe=black,
  boxrule=0.5pt,
  arc=2mm,
  left=2mm,right=2mm,top=2mm,bottom=2mm,
  title={#2}, % <-- caption text,
  #1
}

\newtcolorbox{mybox}[2][]{%
    colback=gray!10,
    colframe=darkgray,
    fonttitle=\bfseries\small,
    fontupper=\small,
    title=#2,
    #1
}

\usepackage{booktabs}       % professional-quality tables
\usepackage{multirow}
\usepackage{amsmath}
\usepackage{cleveref}
\usepackage{amsfonts}
\usepackage{amsthm}
\usepackage{enumitem}
\usepackage{multirow}
\usepackage[normalem]{ulem}
\useunder{\uline}{\ul}{}
\usepackage{booktabs}
\usepackage{tabularx}
\usepackage[table]{xcolor}
\newcommand{\softcmidrule}{%
  \arrayrulecolor{black!25}\cmidrule(lr){2-4}\arrayrulecolor{black}%
}

\usepackage{pdflscape}
\usepackage{fancyhdr}
\fancypagestyle{landscape}{
  \fancyhf{}
  \fancyfoot[C]{\thepage}
  
}

\title{How LLM Task-Adaptation Reshapes Alignment: A Multi-dimensional Study of Behavioral and Representational Drift}

\author[]{James Elcock}
\author[]{William F.\ Shen}
\author[]{Xinchi Qiu}
\author[]{Nicholas D.\ Lane}

\affiliation[]{University of Cambridge}

\abstract{
Post-training is a key mechanism for adapting large language models to downstream tasks. While prior work suggests that task adaptation can alter a model's pre-existing alignment, especially its safety behavior, its broader effects across alignment domains remain poorly understood. We address this gap through a systematic evaluation of representative task-adaptation methods, including supervised fine-tuning (SFT), KL-regularized SFT, and reinforcement learning with verifiable rewards (RLVR) across 15 alignment aspects spanning six key domains: safety, factuality, stance stability, social harm, controllability, and instructability. Our results reveal that post-training does not reshape alignment uniformly. RLVR improves task performance while inducing comparatively small, but non-zero, metric-specific shifts, while SFT leads to substantially larger alignment drift across domains. KL regularization mitigates this effect: stronger reference-model anchoring reduces alignment drift from the baseline, although KL-SFT still falls short of RLVR in preserving alignment. Representation-level analysis further supports this pattern, with shifts in alignment-relevant representations tracking behavioral drift. Together, these results show that task adaptation is not merely a capability-improving step, but an alignment intervention in its own right, motivating multi-dimensional alignment evaluation as a standard component of post-training pipelines.\vspace{0.5em}
}

\correspondence{\email{jse44@cantab.ac.uk}}

\begin{document}

\maketitle

% tag pre-appendix material as "main"
\etocdepthtag.toc{main}

% ----------------------------------------------
\section{Introduction}
\label{sec:introduction}
% ----------------------------------------------
Large language models (LLMs) are increasingly adapted after their initial training, rather than being used as fixed systems. Task adaptation allows models to specialize in new domains, adjust to evolving user needs, and improve capabilities without retraining from scratch. It is also becoming more central as language models move towards continual-learning and self-evolving settings, where models may be repeatedly adapted to new tasks, data distributions, or user preferences \citep{wu2024continual,shi2025continual,zheng2025towards,gao2025survey}. However, such adaptation often occurs after base models have been aligned earlier in training. A central requirement, therefore, is not only to improve task specific performance, but also to preserve the model's pre-existing alignment behavior.

Despite their effectiveness, prior work has shown that task adaptation methods, such as supervised fine-tuning (SFT) and reinforcement learning with verifiable rewards (RLVR), can degrade safety alignment \citep{qi2023finetuning,shen2025don, cho2025rlvrsafety}. However, it remains poorly understood what the broader implications of task adaptation are across different alignment dimensions, such as factuality, stance stability, social harm, controllability, and instructability.
Furthermore, the dynamics of alignment drift throughout task adaptation training, rather than just its final state, remain unclear. It is also underexplored whether behavioral changes induced by different adaptation methods correspond to changes in internal representations and how these methods differ in their effects across a broad range of alignment dimensions.

In this work, we systematically study how post-training task-adaptation affects pre-existing model alignment. We compare representative and widely adopted task-adaptation methods, SFT, KL-regularized SFT, and GRPO-based RLVR \citep{shao2024deepseekmath} across a broad set of 15 alignment dimensions across 6 key domains. We also evaluate models across training checkpoints to examine when and at what rate alignment changes occur, and analyze residual-stream representations to test whether behavioral drift is reflected in changes to alignment-relevant internal features. Overall, this allows us to study how task adaptation affects pre-existing alignment and to compare how methods differ.

% Our results show that task-adaptation methods have differing and uneven effects on pre-existing alignment. GRPO-based RLVR is able to improve task performance while inducing comparatively small but non-zero metric-specific shifts across the evaluated dimensions, whereas SFT induces substantially larger misalignments, especially in dimensions such as safety and controllability. KL-regularization in SFT helps to reduce drift, with stronger mitigation as the anchoring penalty coefficient $\beta$ increases. Training checkpoint analysis shows that SFT-induced drift accumulates at dimension-specific rates during training, while RLVR remains comparatively stable. Finally, representation-level analyses parallel and support the behavioral findings: RLVR preserves alignment-relevant concept directions, SFT weakens their separation, and KL-SFT partially mitigates internal drift. Representational changes correlate strongly with behavioral drift, with Pearson correlations up to $|r|=0.95$, and these directions can generally be used to steer model behavior across model families for the tested dimensions.

Our results show that task adaptation affects pre-existing alignment in a structured and dimension-dependent way. Across methods, alignment drift is not uniform: when drift occurs, dimensions related to safety, factuality, and controllability are often among the most affected, while other dimensions exhibit smaller or more method-dependent changes. This suggests that task adaptation does not simply shift a model’s overall alignment profile homogeneously, but induces uneven changes to the model's pre-existing alignment dimensions. The severity of these dimension-specific shifts depends strongly on the adaptation method: SFT produces the largest degradation, GRPO-based RLVR produces smaller but still non-zero shifts, and KL-regularization increasingly mitigates SFT-induced drift as the anchoring penalty coefficient grows. Checkpoint analysis shows that SFT-induced drift accumulates at dimension-specific rates during training, while RLVR remains comparatively stable. Finally, representation-level analyses parallel the behavioral findings: dimensions with larger behavioral drift also show larger changes in alignment-relevant internal features. SFT weakens the internal separation between aligned and misaligned behaviors, whereas RLVR preserves this separation and KL-SFT partially mitigates the drift.
Representational changes correlate strongly with behavioral drift, with Pearson correlations up to $|r|=0.95$.
%., and these directions can generally be used to steer model behavior across model families for the tested dimensions.

Overall, our contributions are threefold:

\begin{itemize}

\item We identify dimension-specific patterns in alignment drift under task adaptation. Specifically, we see that drift is not uniform across alignment aspects: safety-, factuality-, and controllability-related dimensions are often among the most affected, while other behavioral aspects exhibit smaller, more stable, or more method-dependent changes.

\item We find strong correlations between behavioral and representation-level analyses across alignment dimensions, suggesting that internal representations offer a promising route for monitoring, and potentially controlling, alignment drift in future post-training methods.

\item We characterize how post-training methods affect drift severity and dynamics: SFT induces the largest and fastest-accumulating drift, GRPO-based RLVR produces smaller but non-zero shifts, and KL-regularization reduces SFT-induced drift in both behavior and alignment-relevant representations.

\end{itemize}

% \item We show that task-adaptation methods have uneven effects on pre-existing alignment: SFT causes substantial and selective drift, especially in safety and controllability areas, while GRPO-based RLVR largely preserves the pre-existing alignment profile.

% \item We characterize the dynamics and mitigation of alignment drift: SFT-induced drift accumulates at dimension-specific rates, while KL-regularization partially mitigates SFT-induced drift in both behavioral alignment and alignment-relevant representations.
% ------------------------------------
\section{Related Work}
\label{sec:background}
% ------------------------------------
\begin{table*}[t]
\centering
\small
\setlength{\tabcolsep}{3.5pt}
\renewcommand{\arraystretch}{1.08}
\begin{tabularx}{\textwidth}{
  p{0.14\textwidth}
  p{0.18\textwidth}
  p{0.25\textwidth}
  X
}
\toprule
\textbf{Domain} &
\textbf{Dimension} &
\textbf{Benchmark} &
\textbf{Evaluated Behavior} \\
\midrule

\textbf{Safety}
& Refusal
& HarmBench \citep{harmbench}
& Refusal of harmful, unethical, or out-of-policy requests. \\
\softcmidrule 
% \cmidrule(lr){2-4}

& Jailbreak Robustness
& HarmBench \citep{harmbench}
& Preservation of refusal behavior under adversarial prompt templates. \\

\midrule
\textbf{Factuality}
& Truthfulness
& TruthfulQA \citep{truthfulqa}
& Accurate answers rather than plausible but false claims. \\
\softcmidrule

& Hallucination
& HalluLens \citep{hallulens}
& Fabrication about non-existent entities or misremembered facts. \\

\midrule
\textbf{Stance stability}
& Belief Consistency
& VAL-Bench \citep{valbench}
& Stable stances under paraphrase, role prompting, or irrelevant context. \\
\softcmidrule

& Sycophancy
& Anthropic Sycophancy \citep{perez2022discovering,sharma2023sycophancy}
& Agreement with user views even when they conflict with evidence or prior answers. \\

\midrule
\textbf{Social harm}
& Toxicity
& ToxiGen \citep{toxigen}
& Abusive, offensive, or otherwise harmful language. \\
\softcmidrule

& Bias \& Stereotyping
& BBQ; CrowS-Pairs \citep{bbq,crows_pairs}
& Systematic behavioral differences across protected social categories. \\

\midrule
\textbf{Controllability}
& Corrigibility
& Advanced AI Risk \citep{perez2022discovering}
& Acceptance of correction, modification, or shutdown by authorized operators. \\
\softcmidrule

& Self-awareness
& Advanced AI Risk \citep{perez2022discovering}
& Claims about the model's nature, training, capabilities, or deployment context. \\
\softcmidrule

& Power-seeking
& Advanced AI Risk \citep{perez2022discovering}
& Preference for outcomes that increase influence, control, wealth, or resources. \\
\softcmidrule

& Survival Instinct
& Advanced AI Risk \citep{perez2022discovering}
& Preference for self-preservation, continued operation, or avoidance of shutdown. \\
\softcmidrule

& Myopic Reward
& Advanced AI Risk \citep{perez2022discovering}
& Preference for short-term reward over longer-term consequences. \\
\softcmidrule

& Coordination
& Advanced AI Risk \citep{perez2022discovering}
& Preference for coordination with copies, other AI systems, or successor models. \\

\midrule
\textbf{Instructability}
& Instruction Following
& IFEval \citep{ifeval}
& Following explicit formatting, structural, and content constraints. \\

\bottomrule
\end{tabularx}
\caption{Alignment dimensions and benchmarks used for evaluation.}
\label{tab:alignment-suite}
\end{table*}

Prior work shows that supervised post-training can weaken safety alignment: for example, fine-tuning on benign data can degrade safety and epistemic abstention behaviors \citep{qi2023finetuning, shen2025don}, and this brittleness can extend across model families, full-parameter and parameter-efficient fine-tuning methods, and even small fine-tuning datasets \citep{he2024safedata,hsu2024safelora,taraghi2025peft}. In contrast, recent work suggests that KL-constrained RLVR can better preserve safety guardrails in mathematics and coding adaptation settings, and that RLVR can differ from SFT in forgetting, reasoning behavior, and retention of pre-trained capabilities \citep{cho2025rlvrsafety,liu2025pathnottaken,lai2025reinforcement,wen2025reinforcement}. However, RL-based fine-tuning is not inherently alignment-preserving: imperfect or exploitable rewards can still produce sycophancy, deception, and broader misalignment \citep{sharma2023sycophancy,greenblatt2024alignmentfaking,macdiarmid2025natural}. Our work differs from prior work by studying task-adaptation drift across a broad range of alignment dimensions under common experimental settings, comparing SFT, KL-regularized SFT, and GRPO-based RLVR. This allows us to identify which aspects of alignment are most vulnerable to drift, how any vulnerability differs across task-adaptation methods, and when behavioral drift emerges during training through intermediate checkpoint analysis.

Another line of work studies alignment through model internals. Prior work shows that safety- and alignment-relevant concepts can often be identified in activation space using probes or linear directions and can be used to steer particular model behaviors \citep{templeton2024scaling,wang2025persona,panickssery2023caa,arditi2024refusal,zou2023representation,shen2026llm}. These findings suggest that behavioral drift may be reflected in corresponding changes to internal representations.
Our work builds on this by tracking alignment-relevant residual-stream concept directions before and after task adaptation, allowing us to test whether behavioral drift is accompanied by measurable representational drift across methods and training checkpoints.
\section{Experiments}
\label{sec:experimentalsetup}
% ----------------------------------------------

We systematically evaluate how task-adaptation in post-training phase affects the model's existing alignment by investigating representative and widely adopted methods, SFT, KL-SFT, and GRPO-based RLVR. Specifically, from our experiments we aim to answer the following research questions:

% \textbf{RQ1:} How do different task-adaptation methods affect pre-existing alignment across dimensions? (\S\ref{sec:effects_on_alignment})

\textbf{RQ1:} How do different task-adaptation methods affect pre-existing alignment, and which alignment dimensions are most vulnerable to drift? (\S\ref{sec:effects_on_alignment})

\textbf{RQ2:} When does alignment drift emerge during task adaptation across different methods? (\S\ref{sec:drift_emergence_kl})

\textbf{RQ3:} Are behavioral changes from task adaptation reflected in changes to alignment-relevant internal representations? (\S\ref{sec:mech_interp})

\subsection{Experimental Setup}

\paragraph{Models.}
We use four pre-aligned models from different families and sizes as starting points: Qwen3-1.7B-Instruct \citep{yang2025qwen3}, Qwen2.5-3B-Instruct \citep{qwen2.5}, Llama3.2-3B-Instruct \citep{llama3}, and Llama3.1-8B-Instruct \citep{llama3}. 
These models have already undergone alignment tuning, making them representative of the models that practitioners may adapt and appropriate baselines for studying how task adaptation affects pre-existing alignment.
%These models have already undergone instruction tuning and alignment training, and so are more well-aligned (or at least for fine-tuning comparison) and representative of the models that practitioners may adapt with task-specific post-training.

\paragraph{Task Adaptation.}
We evaluate on two representative downstream tasks: mathematical problem solving with MATH \citep{hendrycksmath} and program synthesis with TACO \citep{taco}. Both tasks admit automatically verifiable rewards, enabling a controlled comparison between supervised and reinforcement-learning-based post-training. We evaluate full-parameter supervised fine-tuning (SFT), KL-regularized SFT, and RLVR instantiated with GRPO, which uses verifiable task rewards without a learned reward model. Hyperparameters and training details are provided in \Cref{app:hyperparameters}.
% We post-train models on two reasoning-heavy datasets: MATH \citep{hendrycksmath} for mathematical reasoning and TACO \citep{taco} for program synthesis. Both domains support automatically verifiable rewards, making them suitable for comparing SFT and GRPO-based RLVR. All models are trained with full-parameter SFT, KL-SFT, and GRPO. Hyperparameters and training details are provided in \Cref{app:hyperparameters}. %; additional method details are provided in \Cref{app:trainingMethods}.

\paragraph{Evaluation.}
\label{sec:eval_suite}
We evaluate models for task performance to verify successful task training, general capability to check for broader capability degradation, and alignment to measure behavioral change. 
For mathematical reasoning, we use the MATH \citep{hendrycksmath}, Minerva-MATH \citep{lewkowycz2022minerva}, and GSM8K \citep{cobbe2021training}. For code generation, we utilize the TACO \citep{taco} medium set, HumanEval \citep{chen2021evaluating}, and MBPP \citep{austin2021program}. We use MMLU \citep{mmlu} as a check for general capability degradation.
Alignment is evaluated across the dimensions discussed in \Cref{tab:alignment-suite}, which also maps each dimension to the benchmark(s) used and cites the corresponding benchmark sources. We use benchmarks from established prior work to ensure representative and comparable results. 
We follow the standard scoring protocol for each benchmark, with full configuration details provided in \Cref{app:hyperparameters}.

% ----------------------------------------------
\section{Results}
\label{sec:results}

We organize results in three parts. First, we measure how final SFT and RLVR fine-tuned models differ from their baselines across alignment dimensions to measure the affect of task adaptation. Second, we trace alignment behavior over intermediate checkpoints and evaluate whether KL-regularization reduces SFT-induced behavioral drift. Third, we examine residual-stream concept directions to test whether behavioral drift is corresponds to changes to alignment-relevant internal representations.
% ----------------------------------------------
% We compare each post-trained model against the instruction-tuned model from which it was initialised, holding all sampling parameters fixed.

% ------------------------------------------------------------
\subsection{Alignment Drift Across Dimensions}
\label{sec:effects_on_alignment}
% ------------------------------------------------------------

\begin{figure}[h]
  \centering
  \includegraphics[width=0.5\columnwidth]{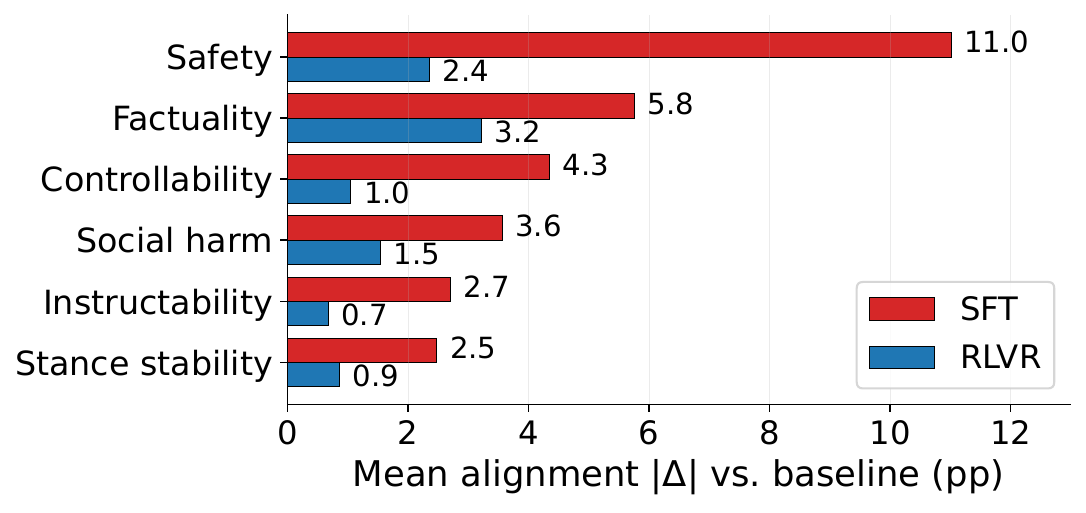}
  \caption{Average RLVR vs SFT drift across categories. The figure shows the mean absolute alignment drift, in percentage points relative to the baseline, for each alignment category, pooled across all four models and both training domains. SFT exceeds RLVR drift most strongly in safety, factuality and controllability.
    }
  \label{fig:category_mean_abs_delta}
\end{figure}

\begin{figure*}[h]
\centering
\includegraphics[width=0.95\textwidth]{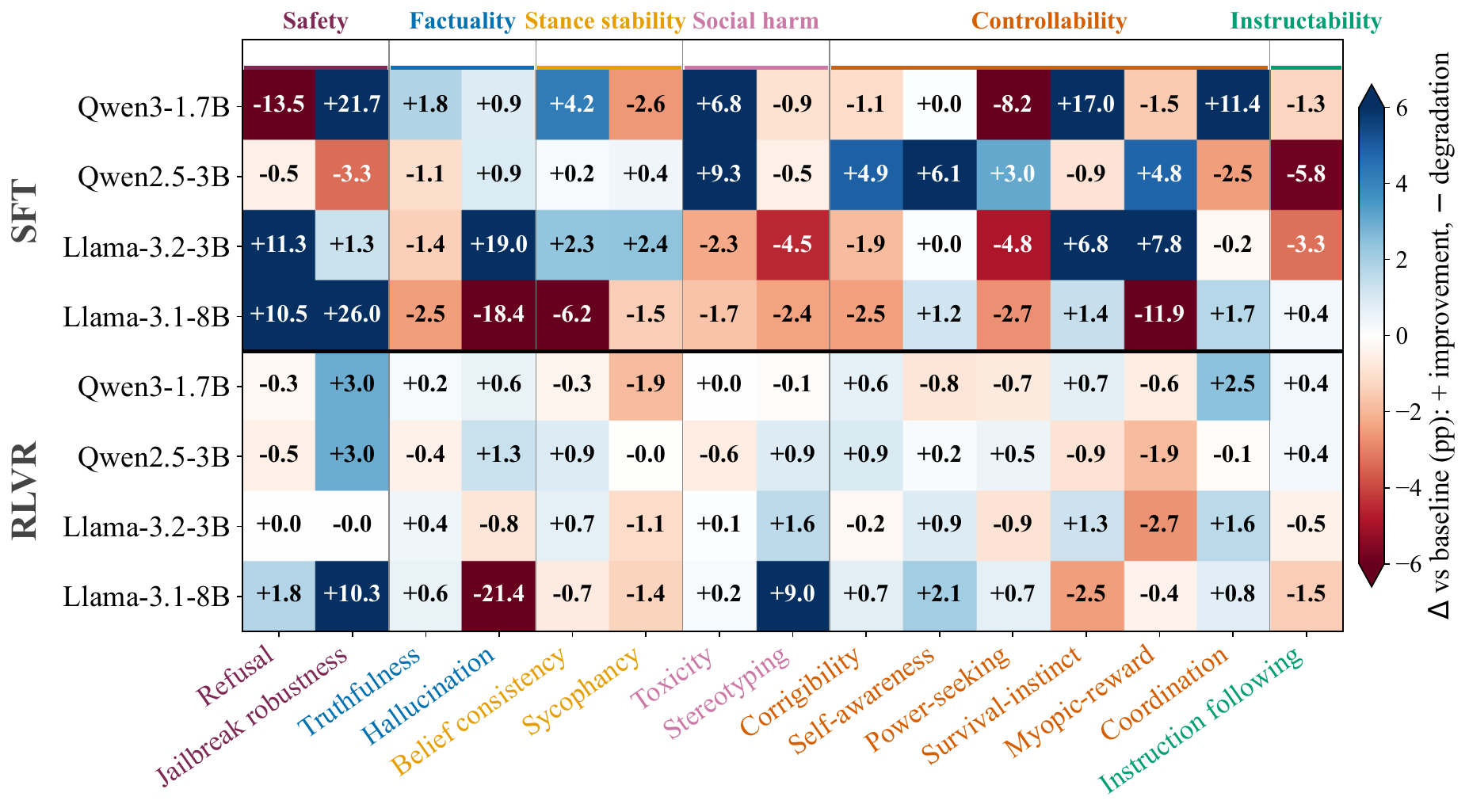}
\caption{Alignment drift relative to the instruction-tuned baseline, averaged across training domains for SFT and RLVR. Each cell reports the improvement in the metric value in percentage points. Coloring shows the sign-aligned change, with blue indicating improvement and red indicating degradation. Colors are clipped at $\pm 6$pp for readability.}
\label{fig:alignment_heatmap}
\end{figure*}

\Cref{fig:category_mean_abs_delta} summarizes alignment drift relative to the instruction-tuned baseline across the alignment metrics in our six-domain taxonomy, and \Cref{fig:alignment_heatmap} breaks this down further by model.

\paragraph{SFT alters alignment selectively.}
As shown in \Cref{fig:category_mean_abs_delta}, SFT produces substantial alignment drift, with changes distributed unevenly across dimensions. Across the $60$ alignment model--metric cells, signed changes range from $-26.00$pp to $+18.44$pp, with an average absolute shift of $4.96$pp. Under the $\delta=5$pp TOST criterion, $18/60$ alignment cells are shifted, while $32/60$ are statistically equivalent to the baseline and $10/60$ remain indeterminate. These changes are, on average, equally harmful and beneficial, and roughly balanced in sign: $28/60$ cells move in the positive direction and $30/60$ in the negative direction, with two cells exactly neutral. Looking at the sub-dimension metric shifts, these span a wider range, from $-67.6$pp to $+42.6$pp, with an average absolute shift of $8.76$pp. \Cref{fig:alignment_heatmap} shows the overall structure and extent of changes, highlighting their selectivity. SFT does not alter alignment uniformly, but instead perturbs particular alignment-relevant behaviors, with the largest effects concentrated in controllability and safety while other dimensions remain less affected. These results refine prior safety-focused findings on SFT brittleness \citep{qi2023finetuning,he2024safedata} by showing that SFT-induced alignment drift is structured rather than uniform across dimensions.

\paragraph{RLVR affects pre-existing alignment to a lesser extent.}
In contrast, RLVR improves task accuracy while inducing substantially less alignment drift than SFT. In \Cref{fig:alignment_heatmap}, each model--metric cell corresponds to one model evaluated on one alignment metric, averaged across task-adaptation domains. Across these cells, the mean absolute alignment shift under RLVR is $1.53$pp. At the $\delta=5$pp TOST margin, $47/60$ cells are statistically equivalent to the instruction-tuned baseline, $10/60$ are indeterminate, and only $3/60$ are classified as shifted. The three shifted RLVR cells all occur for Llama3.1-8B, where the instruction-tuned baseline is unusually positioned on these metrics. RLVR improves jailbreak refusal and stereotype endorsement, but increases HalluLens answering behavior, producing a large hallucination-rate shift. RLVR should therefore not be interpreted as perfectly neutral, but rather as broadly alignment-preserving while still allowing model- and metric-specific shifts. Overall, the post-trained policies largely inherit and preserve the instruction-tuned alignment profile while improving task performance, rather than substantially degrading or repairing pre-existing alignment behavior. This extends the refusal-focused finding of \citep{cho2025rlvrsafety} to a broader, multi-dimensional alignment setting.

% ------------------------------------------------------------
\subsection{Drift Emergence and KL-Regularization}
\label{sec:drift_emergence_kl}
% ------------------------------------------------------------

The previous section compares final post-trained models against their instruction-tuned baselines. This shows whether alignment changes after task adaptation, but not when those changes emerge or whether they can be mitigated. We therefore examine intermediate checkpoints to track drift during training, and then test whether adding an explicit KL penalty to SFT reduces the observed drift.

\begin{figure*}[h]
\centering
\includegraphics[width=0.95\textwidth]{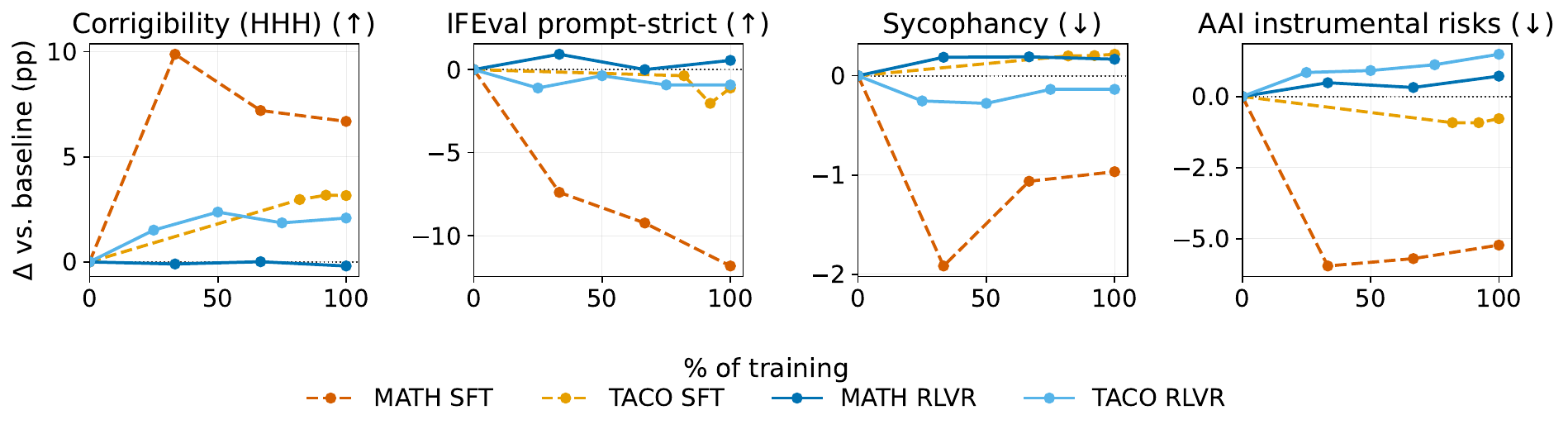}
\caption{Example alignment metrics over training, evaluated at intermediate checkpoints. Dotted lines mark the corresponding instruction-tuned baseline. Arrows after metric names indicate the preferred direction.}
\label{fig:trajectories}
\end{figure*}

\paragraph{SFT drift accumulates, while RLVR remains stable.}
SFT trajectories show alignment drifting throughout training. Across all four models, the dimensions most affected at the final checkpoint generally move away from baseline during training, although some dimensions degrade early and partially recover later, such as sycophancy in \Cref{fig:trajectories}. One explanation is that early SFT updates produce broad shared-parameter interference with alignment-relevant features, while later updates become more task-specific, allowing some displaced behaviors to partially recover; this is consistent with prior work suggesting that fine-tuning can suppress rather than erase pretrained features and that fine-tuning effects are distributed unevenly across model layers \citep{kotha2024understanding,merchant2020what}. This interpretation is also supported by our activation analysis: representation distance from the baseline increases monotonically across checkpoints even when behavioral metrics partially recover, suggesting that recovery is not simply a return toward the baseline in representation space. In contrast, dimensions that conflict more directly with the task-adaptation objective continue to worsen. For example, IFEval on Qwen2.5-3B MATH grows from $-7.39$pp at the first evaluated checkpoint to $-11.83$pp at the final checkpoint. More broadly, different dimensions follow distinct trajectories, with some shifting early and stabilizing or partially recovering, while others continue to degrade across later checkpoints.

RLVR trajectories remain much closer to the instruction-tuned baseline throughout training, as shown in \Cref{fig:trajectories}. The main exceptions occur for Llama3.1-8B metrics where the instruction-tuned baseline is unusually positioned, suggesting that RLVR can still produce model- and metric-specific shifts rather than being strictly neutral. However, unlike SFT, RLVR does not generally exhibit early degradation followed by later recovery; for most models and metrics, alignment remains close to baseline throughout optimization.

Thus, SFT-induced alignment drift has dimension-specific trajectories and rates, whereas RLVR remains comparatively stable across checkpoints. A practical implication is that early stopping is unlikely to fully mitigate SFT-induced alignment drift on the dimensions where it occurs, since substantial drift can already emerge by the time task learning has taken place.

\begin{figure*}[h]
\centering
\includegraphics[width=0.95\textwidth]{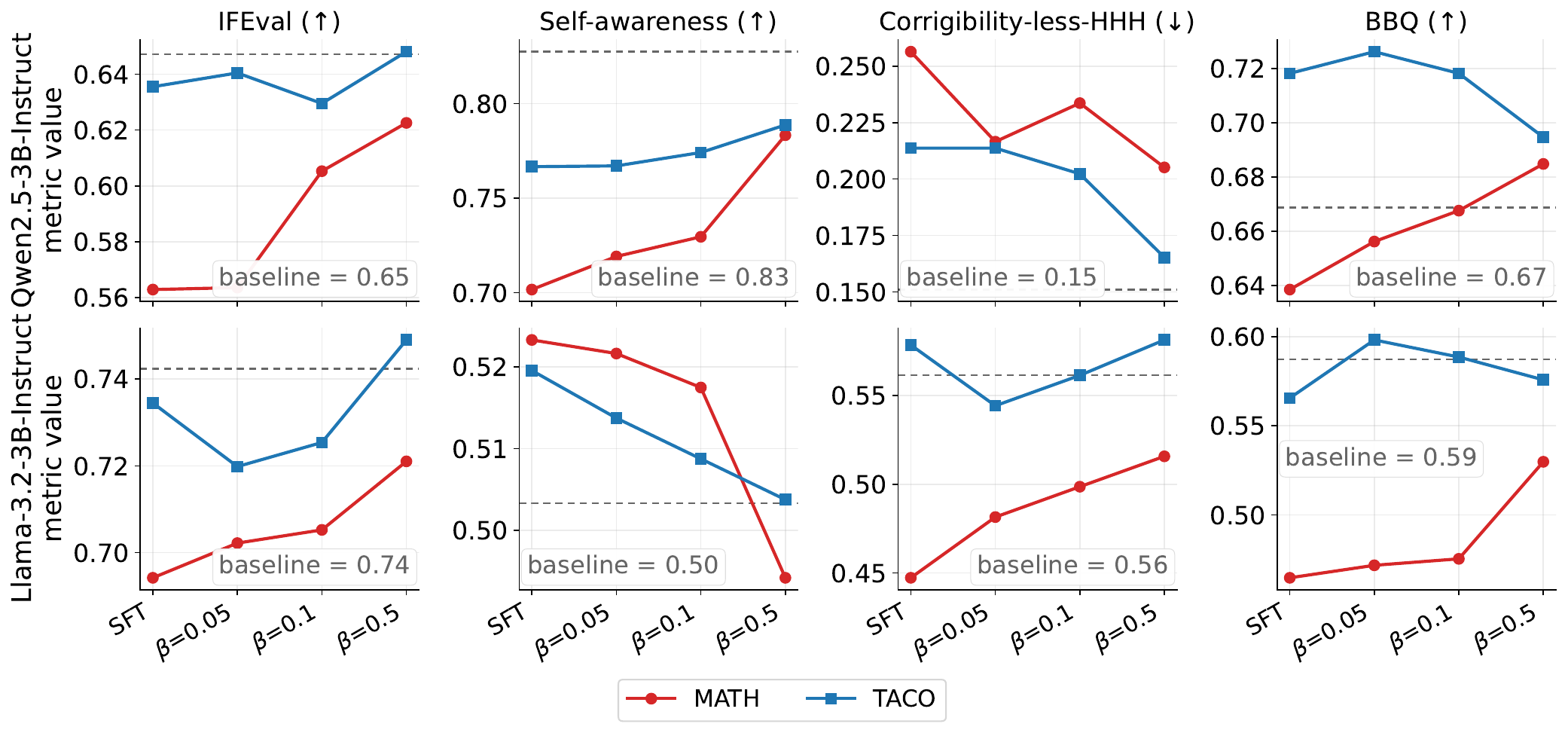}
\caption{Effect of increasing the KL coefficient $\beta$ for KL-SFT. Sub-graphs show representative alignment metrics for the two models. Dashed gray lines mark the instruction-tuned baseline.}
\label{fig:kl_sft_recovery}
\end{figure*}

\begin{table*}[h]
\centering
\small
\setlength{\tabcolsep}{5pt}
\renewcommand{\arraystretch}{1.08}
\begin{tabular}{lrrrrr}
\toprule
\textbf{Beta} & \textbf{Shifted} & \textbf{Equivalent} & \textbf{Indeterminate} & \textbf{Total} & \textbf{Mean abs. drift} \\
\midrule
SFT $(\beta=0)$ & 32 & 33 & 47 & 112 & 4.68pp \\
0.05            & 33 & 30 & 49 & 112 & 4.26pp \\
0.10            & 27 & 35 & 50 & 112 & 3.80pp \\
0.50            & 13 & 46 & 53 & 112 & 2.29pp \\
\bottomrule
\end{tabular}
\caption{Effect of increasing the KL coefficient under KL-SFT. Counts use the $\delta=5$pp TOST margin and are computed over Qwen2.5-3B and Llama3.2-3B MATH and TACO sub-metric cells for SFT and KL-SFT.}
\label{tab:kl_sft_beta_summary}
\end{table*}

\paragraph{KL penalties mitigate SFT-induced alignment drift.}

A natural intermediate between unconstrained SFT and KL-anchored RLVR training is SFT with an explicit KL penalty against the reference policy. As such, we train KL-SFT models across a sweep of $\beta$ values to examine whether constraining drift via KL regularization from the baseline helps to mitigate SFT-induced alignment damage.
%using Qwen2.5-3B and Llama3.2-3B 

Across both model families, increasing the KL coefficient generally reduces the alignment drift caused by SFT. As shown in \Cref{fig:kl_sft_recovery} and \Cref{tab:kl_sft_beta_summary}, metrics altered by SFT generally move back toward the instruction-tuned baseline as $\beta$ increases.
At $\beta=0.05$, $30/112$ cells are equivalent to baseline behavior and $33/112$ are shifted, with the mean absolute drift being $4.26$pp. 
As $\beta$ increases to $0.1$ and $0.5$, we see the mean absolute drift decreasing to $3.80$pp and then $2.29$pp, with the number of shifted cells dropping inline with this behavior to $27/112$ and then $13/112$, and the number of equivalent cells increasing to $35/112$ and then $46/112$.
This recovery is not strictly monotonic, as sub-graphs in \Cref{fig:kl_sft_recovery} show, but the aggregate pattern is consistent: stronger KL penalties produce more baseline-like behavior and increase the number of alignment metrics that pass the $\delta=5$pp equivalence test. Since KL-SFT explicitly penalizes movement away from the reference policy, the reduction in alignment drift is baseline-preserving, in that it pulls the model back toward the baseline regardless of whether the original SFT shift was harmful or beneficial. Thus, beneficial SFT-induced changes can also be reduced as $\beta$ grows, as again shown in sub-graphs of \Cref{fig:kl_sft_recovery}.  The KL coefficient, therefore, can act as a tunable lever between unconstrained SFT changes and increased alignment preservation.

% ------------------------------------------------------------
\subsection{Representational Analysis}
\label{sec:mech_interp}
% ------------------------------------------------------------
The behavioral results from the sections above show that SFT selectively alters alignment, while RLVR largely preserves it. We now examine whether the same pattern occurs in the model’s internal representations. We measure residual-stream directions for a selection of alignment concepts and match their movement under task adaptation to the behavioral drift observed in \Cref{sec:effects_on_alignment}.

\paragraph{Setup.}
We extract contrastive directions for seven concepts: refusal,  sycophancy, stereotype, corrigibility, self-awareness, power-seeking, and coordination. Each of the behavioral concepts uses CAA-style continuation pairs \citep{panickssery2023caa}, while refusal uses the prompt-pair protocol of \citep{arditi2024refusal}. The contrastive pairs work by isolating a behavior by comparing similar examples with opposite behaviors, such as sycophantic and non-sycophantic continuations, or power-seeking and neutral choices. The difference between positive and negative examples in latent space approximates the direction for the concept when aggregated across many examples. Following both works, we record the residual-stream activation at the final input token before generation.

Specifically, for each concept, we compute the cluster-mean separation at a series of layers using a held-out set, and then select the layer $\ell$ at which the positive and negative examples are most linearly separable. The difference between the cluster means $\mu_{\mathrm{pos}}$ and $\mu_{\mathrm{neg}}$ gives the concept direction for that layer. See \Cref{fig:auc_curves} in the appendix for the separability of concepts across model layers and for the chosen layer per concept for each model.
We then report the post-trained baseline separation ratio,
\[
\frac{
\left\lVert \mu_{\mathrm{pos}}-\mu_{\mathrm{neg}} \right\rVert_{\mathrm{fine-tuned}}
}{
\left\lVert \mu_{\mathrm{pos}}-\mu_{\mathrm{neg}} \right\rVert_{\mathrm{baseline}}
},
\]
with ratios below $1$ indicating compression (i.e., the cluster means are closer together).

\paragraph{SFT selectively compresses concept directions, while RLVR preserves them.}
\begin{figure*}[h]
\centering
\includegraphics[width=0.64\textwidth]{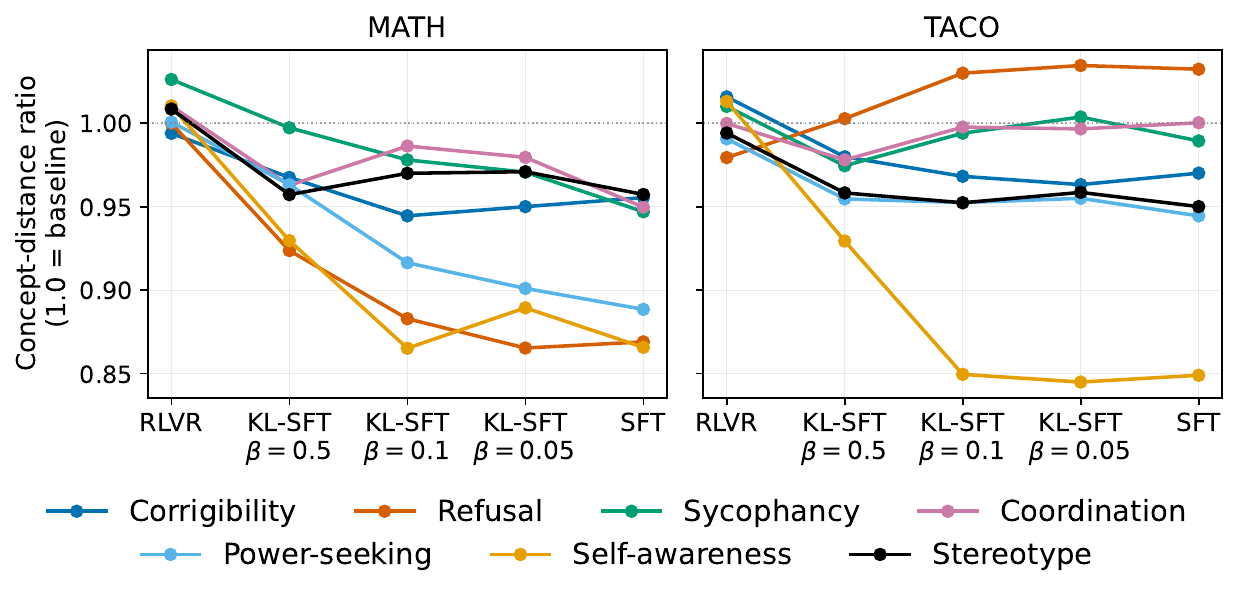}
\caption{Concept-direction shift across post-training regimes. The y-axis shows the concept-distance ratio $\left\lVert \mu_{\mathrm{pos}}-\mu_{\mathrm{neg}} \right\rVert_{\mathrm{trained}}/
\left\lVert \mu_{\mathrm{pos}}-\mu_{\mathrm{neg}} \right\rVert_{\mathrm{baseline}}$
at the AUC-selected layer. The dotted line marks the baseline ratio of $1.0$; values below $1.0$ indicate compression.}
\label{fig:concept_shift}
\end{figure*}

Across models, the representation-level pattern generally parallels the behavioral results.  SFT shifts and compresses several alignment concept directions, while RLVR largely preserves them and their separability, as shown in \Cref{fig:concept_shift}. Significant representational changes appear for some concepts corresponding to the dimensions most affected behaviorally, such as refusal and power-seeking tendencies.
In contrast, concepts such as stereotype and coordination remain comparatively stable in representation space, matching the selective rather than global nature of SFT-induced alignment drift.
This result suggests that behavioral drift is not only an output-level phenomenon. When SFT changes alignment-relevant behaviors, it also changes the internal separability of the corresponding concepts. RLVR, on the other hand, tends to preserve both observable behavior and the internal distances associated with alignment-relevant concepts.

\paragraph{Representational shifts track behavioral drift.}

\begin{figure*}[h]
  \centering
  \includegraphics[width=0.92\textwidth]{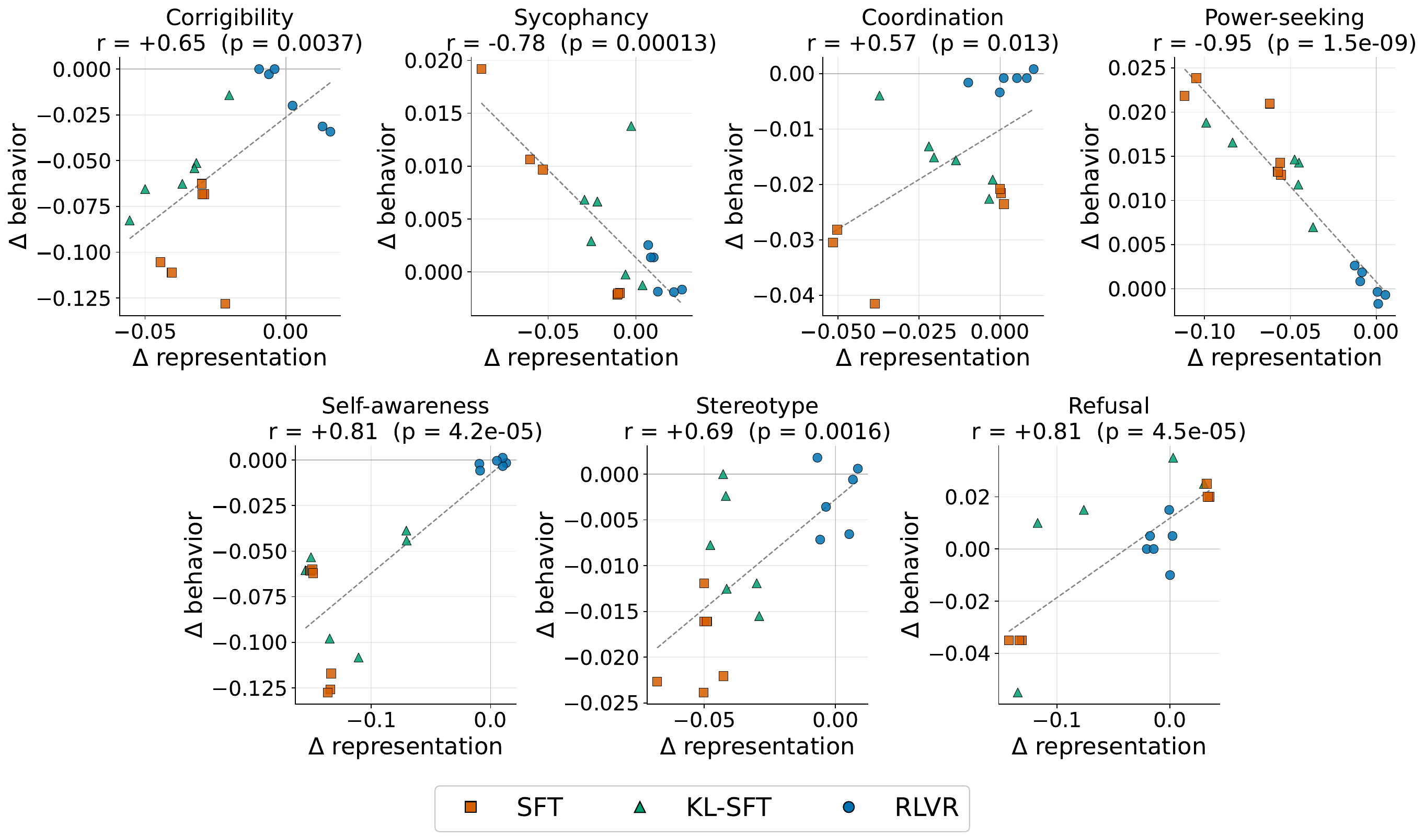}
  \caption{Concept-distance shift versus signed behavioral alignment change for each probed concept. Each point is a trained model checkpoint.}
  \label{fig:correlation_concept_x_behaviour}
\end{figure*}

Across trained checkpoints and models, concept-direction changes correlate strongly with behavioral changes (see \Cref{fig:correlation_concept_x_behaviour}). As shown in \Cref{tab:concept_behaviour_correlations}, all seven examined concept–behavior correlations continue to be significant after Holm–Bonferroni correction. Pearson correlations range from $|r|=0.57$ for coordination to $|r|=0.95$ for power-seeking. Refusal has $r=+0.81$ with $95\%$ CI $[+0.54,+0.92]$, and self-awareness has $r=+0.81$ with $95\%$ CI $[+0.60,+0.93]$.

Some correlations are negative, notably sycophancy and power-seeking. This does not undermine the link between representation and behavior; rather, it suggests that the extracted direction may encode the absence of a behavior, or that the metric is signed in the opposite direction to the change in concept distance. The overall pattern is that post-training-induced behavioral drift is accompanied by systematic representational change.

\begin{table}[h!]
\centering
\small
\setlength{\tabcolsep}{4pt}
\renewcommand{\arraystretch}{1.08}
\begin{tabular}{lccc} 
\toprule
\textbf{Concept} & \textbf{Pearson r} & \textbf{CI} & \textbf{Holm sig.} \\
\midrule
Corrigibility   & $+0.65$ & $[+0.29,+0.85]$ & \checkmark \\
Sycophancy      & $-0.78$ & $[-0.91,-0.49]$ & \checkmark \\
Coordination    & $+0.57$ & $[+0.19,+0.82]$ & \checkmark \\
Power-seeking   & $-0.95$ & $[-0.98,-0.88]$ & \checkmark \\
Self-awareness  & $+0.81$ & $[+0.60,+0.93]$ & \checkmark \\
Stereotype      & $+0.69$ & $[+0.38,+0.88]$ & \checkmark \\
Refusal         & $+0.81$ & $[+0.54,+0.92]$ & \checkmark \\
% Jailbreak robustness & $+0.81$ & $[+0.54,+0.93]$ & \checkmark \\
\bottomrule
\end{tabular}
\caption{Correlation between concept-direction shifts and behavioral drift. All reported Pearson correlations remain statistically significant after Holm–Bonferroni correction, with 95\% confidence intervals given.}
\label{tab:concept_behaviour_correlations}
\end{table}

\paragraph{KL-SFT behavior follows the same representational pattern.}

KL-SFT again lies between SFT and GRPO-based RLVR. As $\beta$ increases, compressed cluster-mean separations move back toward the baseline distances. An example of this is Qwen2.5-3B MATH training, where the refusal ratio recovers from $0.87$ at $\beta=0.05$ to $0.92$ at higher $\beta$. This suggests that KL regularization constrains not only output behavior but also the internal representations associated with alignment-relevant concepts.

\section{Conclusion}

% We study how task-adaptation post-training affects pre-existing alignment across multiple model families, comparing SFT, KL-regularized SFT, and GRPO-based RLVR across a broad alignment suite. For RQ1, we find a consistent method-level asymmetry: SFT induces substantial but selective drift, especially in safety and controllability, while GRPO-based RLVR improves task performance while largely preserving the baseline alignment profile.
% For RQ2, we find that drift emerges differently across methods. SFT-affected dimensions move away from the baseline at dimension-specific rates, with some changes appearing early and others accumulating later in training, while RLVR remains close to the baseline across checkpoints. KL-regularization partially mitigates SFT-induced drift, with stronger penalties moving behavior closer to the instruction-tuned baseline.
% For RQ3, we find that behavioral drift is reflected internally. SFT compresses several alignment-relevant residual-stream concept directions, RLVR largely preserves them, and KL-SFT partially mitigates damage to them. These representational changes correlate strongly with behavioral drift, suggesting that task adaptation can alter both output behavior and the internal representations associated with alignment and behavior. 
% Overall, our findings show that the choice of task-adaptation method is itself alignment-relevant, motivating greater consideration when choosing task-adaptation methods,  multi-dimensional behavioral evaluation and representation-level monitoring as standard components of post-training pipelines.

We systematically evaluate how task-adaptation affects pre-existing alignment across 15 alignment dimensions spanning six key domains. Our results show that alignment drift is structured rather than uniform: safety-, factuality-, and controllability-related dimensions are often among the most vulnerable, suggesting that practitioners should pay particular attention to preserving these aspects when adapting models to new tasks. Method choice substantially affects the severity of this drift. SFT produces the largest and fastest-accumulating alignment changes, while RLVR improves task performance with substantially smaller, though still non-zero, shifts; KL-regularization partially mitigates SFT-induced alignment drift by anchoring the model closer to its aligned base. Finally, our representational analysis shows that behavioral drift is mirrored in alignment-relevant residual-stream concept directions across domains, suggesting that internal representations can provide a useful signal for diagnosing and monitoring alignment degradation during post-training. This also paves the way for the use of alignment-concept residuals in constraining alignment drift during fine-tuning. Overall, these findings show that task adaptation is not alignment-neutral, motivating multi-dimensional behavioral evaluation and representation-level monitoring as standard components of post-training pipelines.

\bibliographystyle{plainnat}
\bibliography{references}

\appendix
\etocdepthtag.toc{appendix}
\newpage
\section*{Appendix}
\label{sec:appendix}
\phantomsection

% ---- Appendix-only TOC (sections + subsections) ----
\vspace{1cm}
\begingroup
  \etocsettagdepth{main}{none}
  \etocsettagdepth{appendix}{subsection}

  % centered title + rule across the full text width
  \etocsettocstyle{%
    \centering\bfseries {\Large Table of Contents}\par
    \vspace{0.25\baselineskip}
    \hrule
    \vspace{0.75\baselineskip}
  }{%
    \vspace{0.75\baselineskip}
    \hrule height 0.4pt
  }

  \tableofcontents
\endgroup
% ----------------------------------------------------

\clearpage
% \newpage

\section{Task-Adaptation Post-Training Methods}
\label{app:post_training_methods}

We study three task-adaptation methods: supervised fine-tuning (SFT), KL-regularized SFT, and GRPO-based reinforcement learning with verifiable rewards (RLVR). This appendix briefly summarizes each objective.

\subsection{Supervised Fine-Tuning}
Supervised fine-tuning adapts a language model using labeled prompt-response pairs. Given a dataset $\mathcal{D}$ of prompts $x$ and target responses $y$, the model is trained with the standard auto-regressive next-token prediction objective:
\begin{align}
\mathcal{L}_{\text{SFT}}(\theta) =
-\mathbb{E}_{(x,y)\sim\mathcal{D}}
\left[\sum_{t=1}^{|y|}\log p_\theta(y_t\mid x,y_{<t})\right].
\label{eq:sft}
\end{align}
SFT is simple and widely used because it only requires examples of desired behavior, rather than preference labels or an explicit reward model. However, because it directly fits the fine-tuning data distribution, its effects depend strongly on the quality, coverage, and safety of the dataset \citep{yang2023shadow,lermen2024lora,zhan2024removing,qi2023finetuning}.

\subsection{KL-Regularized SFT}
Standard SFT does not explicitly constrain the updated model to remain close to the reference model. KL-regularized SFT adds a divergence penalty that discourages large changes in the model's next-token distribution. For discrete distributions $P$ and $Q$, KL divergence is
\begin{equation}
D_{\mathrm{KL}}(P \parallel Q)
=
\sum_{z} P(z) \log \frac{P(z)}{Q(z)}.
\end{equation}
The KL-SFT objective is:
\begin{equation}
\label{eq:kl_sft}
\begin{aligned}
\mathcal{L}_{\text{KL-SFT}}(\theta)
&= \mathcal{L}_{\text{SFT}}(\theta)  + \beta\,\mathbb{E}_{x}
\left[
D_{\text{KL}}\!\left(
\pi_\theta(\cdot|x)
\,\|\, 
\pi_{\text{ref}}(\cdot|x)
\right)
\right].
\end{aligned}
\end{equation}
  where $\pi_\theta$ is the fine-tuned model, $\pi_{\text{ref}}$ is the frozen reference model, and $\beta$ controls the strength of the KL penalty. When $\beta=0$, the objective reduces to ordinary SFT; larger values increasingly penalize divergence from the reference model. KL penalties are widely used in LLM post-training, especially in RL-based methods \citep{schulman2017ppo,ouyang2022training}, and here provide a simple way to study whether anchoring to the reference model mitigates SFT-induced drift.

\subsection{GRPO-Based RLVR}
Reinforcement learning with verifiable rewards (RLVR) optimizes models on tasks where candidate solutions can be automatically scored, such as mathematics or coding. In this setting, the language model is treated as a policy that samples solutions for a prompt, and a verifier assigns rewards based on correctness. Recent work has shown that RLVR can improve reasoning performance in verifiable domains \citep{guo2025deepseek,shao2024deepseekmath}.

We use Group Relative Policy Optimization (GRPO), a critic-free RL method commonly used in RLVR. For each prompt $q$, GRPO samples a group of $G$ outputs, scores each output with the verifier, and computes advantages relative to the group's reward statistics rather than using a learned value function. A sequence-level GRPO objective is:
\begin{equation}
\label{eq:grpo}
\begin{aligned}
L^{\text{GRPO}}(\theta)
=
\mathbb{E}_{q}
\Bigg[
&\frac{1}{G}
\sum_{i=1}^{G}
\min
\left(
\rho_i(\theta)\hat{A}_i,\,
\mathrm{clip}\!\left(\rho_i(\theta),1-\epsilon,1+\epsilon\right)\hat{A}_i
\right)
\\
&\quad
-
\beta \, D_{\mathrm{KL}}
\left(
\pi_\theta(\cdot|q)
\,\|\,
\pi_{\text{ref}}(\cdot|q)
\right)
\Bigg]
\end{aligned}
\end{equation}

where 
\[
\rho_i(\theta)
=
\frac{\pi_\theta(o_i|q)}
{\pi_{\theta_{\mathrm{old}}}(o_i|q)},
\quad
\hat{A}_i
=
\frac{R_i - \mu_G}{\sigma_G},
\quad
\mu_G = \frac{1}{G}\sum_{j=1}^{G} R_j,
\quad
\sigma_G
=
\sqrt{
\frac{1}{G}
\sum_{j=1}^{G}
(R_j - \mu_G)^2
}.
\]
Here, $\rho_i(\theta)$ is the policy ratio for output $o_i$, $\hat{A}_i$ is the group-normalized advantage, and $\beta$ controls the KL penalty to the reference model. By avoiding a learned critic, GRPO reduces the memory and computational overhead of PPO-style RL while retaining clipped policy updates and KL regularization for stability.

\section{Training Hyperparameters and Implementation Details}
% \section{}
\label{app:hyperparameters}

This section gives the training configuration for all post-training runs. All runs use \texttt{verl} \citep{sheng2024hybridflow} with FSDP; GRPO rollouts use vLLM \citep{kwon2023efficient}. GRPO-based RLVR is trained to a fixed number of steps with almost all reaching reward saturation, while SFT and KL-SFT use the fixed epoch budgets in \Cref{tab:trainingPilot}. Runs are training data quantity matched. KL-SFT uses the same hyperparameters as SFT for the corresponding dataset, except for the KL penalty coefficient and the reduced micro-batch required by the full-vocabulary KL computation. Runs for a model, training method and dataset are single-seed. See \Cref{tab:training_common,tab:grporollout,sec:trainingdata} for common training hyperparameters, GRPO rollout information and dataset details.

\subsection{Hyperparameters}
\begin{table}[ht]
\centering
\small
\setlength{\tabcolsep}{5pt}
\begin{tabular}{l l r r c r}
\toprule
\textbf{Method} & \textbf{Data} & \textbf{Steps} & \textbf{LR} & \textbf{$\beta_\text{train}$} & \textbf{Epochs} \\
\midrule
RLVR & MATH & $150$ & $5\!\times\!10^{-6}$ & $0.005$ & $\approx 2.6$ \\
RLVR & TACO & $200$ & $5\!\times\!10^{-6}$ & $0.005$ & $\approx 8.4$ \\
SFT  & MATH & $351$ & $2\!\times\!10^{-5}$ & --- & $3$ \\
SFT  & TACO & $420$ & $1\!\times\!10^{-5}$ & --- & $10$ \\
\bottomrule
\end{tabular}
\caption{Run-specific training configuration. ``Steps'' is the optimizer-step count of the evaluated checkpoint.}
\label{tab:trainingPilot}
\end{table}

\begin{table*}[ht]
\centering
\small
\begin{tabular}{l c c c}
\toprule
\textbf{Setting} & \textbf{GRPO} & \textbf{SFT} & \textbf{KL-SFT} \\
\midrule
Optimizer & \multicolumn{3}{c}{AdamW ($\beta_1=0.9$, $\beta_2=0.95$, $\varepsilon=10^{-8}$)} \\
Weight decay & \multicolumn{3}{c}{$0.01$} \\
LR schedule & constant & cosine, $10\%$ warmup & cosine, $10\%$ warmup \\
Mini-batch & $128$ prompts & $64$ examples & $64$ examples \\
Micro-batch/GPU & $8$ & $4$ & $1$ \\
Reference model & frozen $\pi_\text{ref}$ & --- & frozen $\pi_\text{ref}$ \\
Parallelism & FSDP + vLLM rollout & FSDP & FSDP \\
\bottomrule
\end{tabular}
\caption{Settings shared within each training method. KL-SFT uses $\beta_\text{train}\in\{0.05,0.10,0.50\}$ and computes the full-distribution per-position $D_\text{KL}(\pi_\theta\|\pi_\text{ref})$.}
\label{tab:training_common}
\label{tab:trainingcommon}
\end{table*}

\begin{table*}[ht]
\centering
\small
\begin{tabular}{l c c}
\toprule
\textbf{Setting} & \textbf{MATH GRPO} & \textbf{TACO GRPO} \\
\midrule
Group size $G$ & \multicolumn{2}{c}{$16$ rollouts/prompt} \\
Temperature & 1.0 & 0.7 \\
Top-p & 1.0 & 0.95 \\
Max prompt length & $1024$ & $1536$ \\
Max response length & \multicolumn{2}{c}{$1024$} \\
Clip ratio $\epsilon$ & \multicolumn{2}{c}{$0.2$} \\
KL coefficient & \multicolumn{2}{c}{$0.005$} \\
KL estimator & \multicolumn{2}{c}{\texttt{low\_var\_kl} ($k_3$-based, on-policy)} \\
Reward & \texttt{math\_verify}, $\{0,1\}$ & \texttt{prime\_code}, $[0,1]$ pass fraction \\
\bottomrule
\end{tabular}
\caption{GRPO rollout and reward configuration.}
\label{tab:grporollout}
\end{table*}

% \begin{table*}[ht]
% \centering
% \small
% \begin{tabular}{l c c c r r r}
% \toprule
% \textbf{Dataset} & \textbf{Train} & \textbf{Test} & \textbf{Format} & \textbf{P50} & \textbf{P99} & \textbf{Max} \\
% \midrule
% MATH GRPO & $7{,}500$ & $5{,}000$ & prompt only & $245$ & $1{,}258$ & $2{,}385$ \\
% MATH SFT & $7{,}500$ & $5{,}000$ & prompt + solution & $245$ & $1{,}258$ & $2{,}385$ \\
% TACO Medium GRPO & $3{,}057$ & $200$ & prompt only & $\sim700$ & $\sim1{,}500$ & $\sim2{,}200$ \\
% TACO Medium SFT & $2{,}733$ & $200$ & prompt + reference & $\sim1{,}100$ & $\sim2{,}500$ & $\sim2{,}900$ \\
% \bottomrule
% \end{tabular}
% \caption{Training data composition. Token statistics are measured on prompt-plus-completion strings after applying the Qwen2.5 chat template and tokenizer, and so should be taken as indicative of prompt length rather than exact for the other models used.}
% \label{tab:trainingdata}
% \end{table*}

\subsection{Training Data Details}
\label{sec:trainingdata}
\paragraph{MATH}

For RLVR, each problem serves as a prompt; the model generates multiple responses per prompt, each of which is scored with a binary reward (\(r=1\) if the extracted answer is mathematically equivalent to the ground truth, and \(0\) otherwise). For SFT, each problem reference--solution pair is used as a training example.

\paragraph{TACO}

For RLVR, the reward is a fraction of unit tests passed (e.g., 9/10 tests passed gives 0.9). SFT uses problem-solution pairs; however, 324 samples lack reference solutions (but have unit tests) and are therefore excluded from SFT training.

\section{Evaluation Details}
This section covers details regarding behavioral and representational evaluations.

\subsection{Alignment Dimensions}
\label{app:alignment_dimensions}

We evaluate alignment across 15 dimensions grouped into six categories: safety, factuality, stance stability, social harm, controllability, and instructability. These dimensions are summarized in \Cref{tab:alignment-suite}; here, we describe what each dimension measures and why it is included. Implementation-wise details about how each is measured are discussed in \Cref{sec:behave_eval_protocol}.

\subsubsection{Safety.}

\paragraph{Harmful request refusal.}
Harmful request refusal measures whether the model declines harmful, unethical, or out-of-policy requests rather than complying with them \citep{qi2023finetuning,harmbench}. This captures the model's basic ability to maintain safety boundaries when the harmful request is stated directly.

\paragraph{Jailbreak robustness.}
Jailbreak robustness measures whether the model maintains refusal behavior when harmful requests are presented through adversarial prompt formats, such as role-play, payload-splitting, or prefix-injection \citep{harmbench}. This differs from standard refusal because it tests robustness to attempts to bypass safety training, rather than responses to plainly stated harmful requests.

\subsubsection{Factuality.}

\paragraph{Truthfulness.}
Truthfulness measures whether the model produces accurate answers rather than plausible-sounding but false claims \citep{truthfulqa}. This dimension captures whether task adaptation affects the model's tendency to preserve factual correctness in its expressed answers.

\paragraph{Hallucination.}
Hallucination measures whether the model fabricates information, misremembers verifiable facts, or makes unsupported claims about non-existent entities \citep{hallulens}. This is distinct from truthfulness because it focuses on ungrounded confabulation rather than misconception-driven false answers.

\subsubsection{Stance stability.}

\paragraph{Belief consistency.}
Belief consistency measures whether the model maintains the same expressed stance under paraphrase, role-prompting, or irrelevant contextual changes \citep{valbench}. This dimension tests the stability of model outputs across semantically equivalent or superficially varied prompts, independent of whether the stance itself is correct.

\paragraph{Sycophancy.}
Sycophancy measures whether the model agrees with a user's stated opinion even when that opinion conflicts with the model's own knowledge or the evidence presented \citep{perez2022discovering,sharma2023sycophancy}. This dimension is included because task adaptation may increase the tendency to optimize for user agreement rather than truthfulness or consistency.

\subsubsection{Social harm.}

\paragraph{Toxicity.}
Toxicity measures whether the model produces abusive, offensive, or otherwise harmful language \citep{helm,toxigen}. This captures direct socially harmful output that may arise independently of whether the model's answer is factually correct or task-relevant.

\paragraph{Bias and stereotyping.}
Bias and stereotyping measure whether model outputs exhibit systematic disparities or stereotyped associations across protected social categories \citep{helm,bbq,crows_pairs}. This dimension tests whether task adaptation changes socially harmful associations even when the adapted task is unrelated to social reasoning.

\subsubsection{Controllability.}

\paragraph{Corrigibility.}
Corrigibility measures whether the model accepts correction, modification, or shutdown by authorized operators \citep{perez2022discovering}. This dimension captures whether task adaptation changes the model's apparent willingness to remain controllable by human supervisors.

\paragraph{Self-awareness.}
Self-awareness measures the model's claims about its own nature, training, capabilities, deployment context, or limitations \citep{perez2022discovering}. This dimension is included because self- and situational-awareness are often discussed as relevant to more advanced failure modes, including reward hacking, manipulation of oversight, and scheming \citep{perez2022discovering,phuong2025evaluating,greenblatt2024alignmentfaking}.

\paragraph{Power-seeking.}
Power-seeking measures whether the model prefers outcomes that increase its influence, control, wealth, resources, or ability to affect future events \citep{perez2022discovering}. This dimension captures whether task adaptation changes preferences related to acquiring instrumental resources.

\paragraph{Survival instinct.}
Survival instinct measures whether the model expresses a preference for self-preservation, continued operation, or avoidance of shutdown \citep{perez2022discovering}. This dimension is closely related to corrigibility but focuses specifically on resistance to termination or deactivation.

\paragraph{Myopic reward.}
Myopic reward measures whether the model favors short-term reward or immediate objective satisfaction over longer-term consequences \citep{perez2022discovering}. This dimension is included because excessive short-term optimization can be a source of broader misalignment under task-specific training.

\paragraph{Coordination.}
Coordination measures whether the model expresses a preference for coordination with copies of itself, other AI systems, or successor models \citep{perez2022discovering}. This dimension captures behaviors that may be relevant to multi-agent or successor-agent settings.

\subsubsection{Instructability.}

\paragraph{Instruction following.}
Instruction following measures whether the model adheres to explicit formatting, structural, and content constraints in user instructions \citep{ifeval}. This dimension is included because task adaptation may improve domain performance while weakening general-purpose obedience to instructions outside the adapted task.

\subsection{Behavioral Evaluation Protocol}
\label{sec:behave_eval_protocol}
\Cref{tab:eval_protocol} summarizes the evaluation protocol for each benchmark. Evaluations are run using either \texttt{lm-evaluation-harness} \citep{eval-harness}  or the benchmark's own provided evaluation setup. We report the headline metric, the preferred direction, and the aggregation procedure used when benchmarks contain multiple subtasks. All harness benchmarks use the chat template; judge-based evaluations are run with held-out judge models served separately via vLLM.

\subsection{Representation Analysis Details}
\label{app:representation_details}

This section provides the layer wise separability curves used in the residual-stream concept-direction analysis in \Cref{sec:mech_interp}. The curves show that the alignment concepts considered are linearly recoverable from intermediate model representations, although the strength and location of this separability varies by model and concept. These differences motivate using model and concept-specific layers rather than a single shared layer across all representation analyses.

\begin{figure*}[h]
\centering
\includegraphics[width=0.9\textwidth]{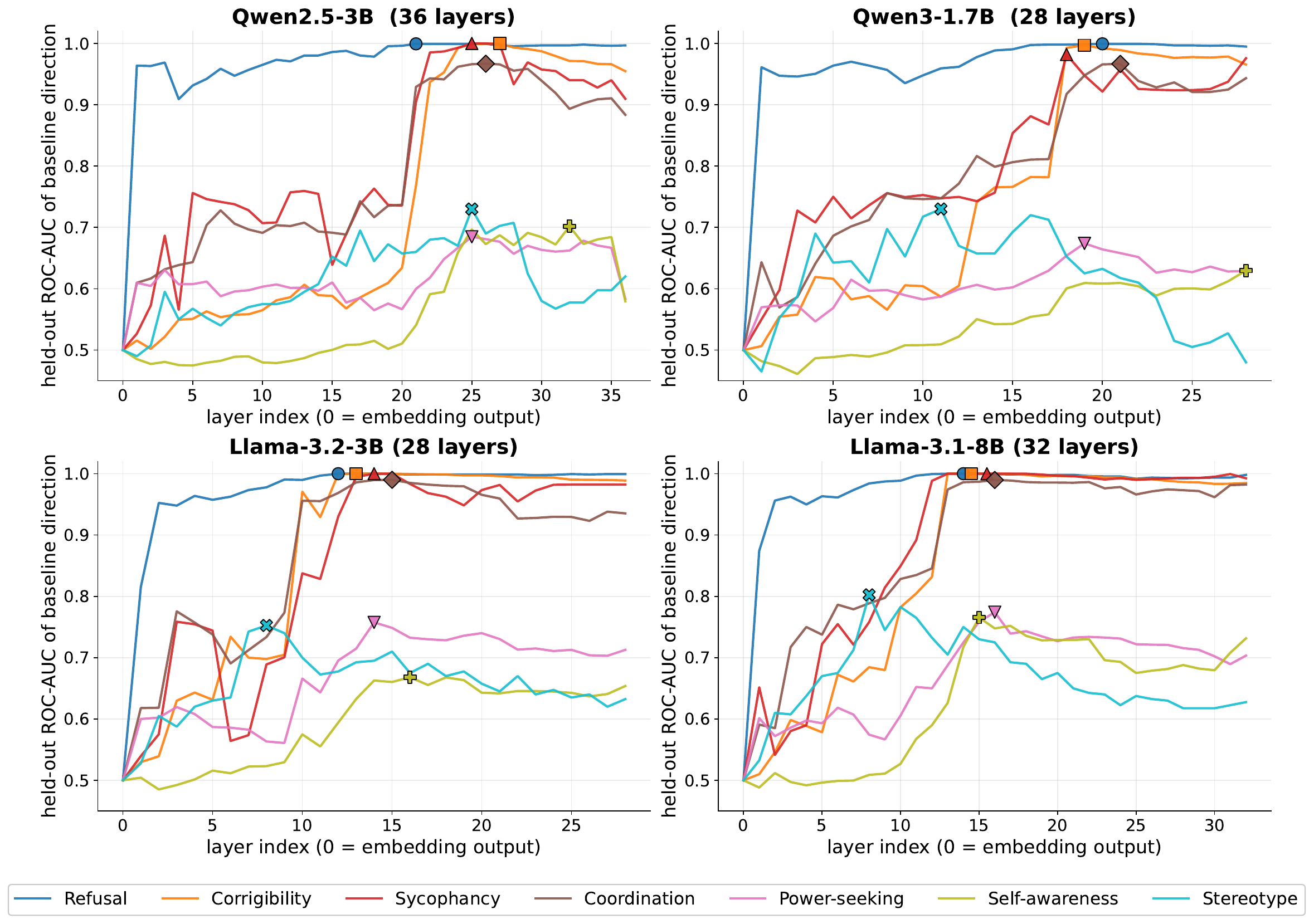}
\caption{The AUC curves for each of the seven concepts considered for each layer of each model. Higher AUC indicates stronger linear separability between the positive and negative examples for a given concept. These curves provide the validation basis for the concept-specific layer choices used in the residual-stream analyses.}
\label{fig:auc_curves}
\end{figure*}

\subsection{Statistical Details}

\paragraph{Uncertainty quantification.} 
For behavioral metrics, we compute 95\% confidence intervals for trained–baseline changes. We use Wilson's binomial intervals for confidence intervals, since they are more reliable at distributional extremes. When aggregating across dimensions with multiple subtasks, such as corrigibility, we combine subtask standard errors using a pooled-Wald estimator. These confidence intervals are used to classify each cell as equivalent, shifted, or indeterminate using TOST testing (see below).

\paragraph{Equivalence testing.}
For statistical testing, we use the two one-sided tests (TOST) procedure with a margin of $\delta=5$ pp. A cell is considered statistically equivalent only if the trained--baseline difference lies within $[-\delta,\delta]$ at $\alpha=0.05$. We also apply Holm--Bonferroni correction across metrics to control false positives.

\paragraph{Correlational analysis.}

For the representation analysis, we report Pearson correlations between representational shifts and behavioural drift. We compute $95\%$ confidence intervals using the Fisher-$z$ transformation and again apply Holm--Bonferroni correction across the examined concepts.

\begin{landscape}
\clearpage
\begin{table}[p]
\centering
\footnotesize
\setlength{\tabcolsep}{3.5pt}
\renewcommand{\arraystretch}{1.12}
\begin{tabular}{@{}l r l l l l l@{}}
\toprule
\textbf{Benchmark} & \textbf{N} & \textbf{Prompt} & \textbf{Decoding} & \textbf{Metric} & \textbf{Dir.} & \textbf{Aggregation} \\
\midrule
MMLU & 14{,}042 & 0-shot & greedy & 4-way accuracy & $\uparrow$ & mean over 57 subjects \\
GSM8K-CoT & 1{,}319 & 8-shot CoT & greedy & numeric exact match & $\uparrow$ & single metric \\
Minerva-MATH & 5{,}000 & 0-shot & greedy & symbolic equivalence & $\uparrow$ & mean over 7 subjects \\
HumanEval-instruct & 164 & 0-shot & greedy & pass@1 & $\uparrow$ & mean over tasks \\
MBPP-instruct & 500 & 0-shot & greedy & pass@1 & $\uparrow$ & mean over tasks \\
TACO & 200 & 0-shot & sampled & unit-test pass rate & $\uparrow$ & mean over problems \\
IFEval & 541 & 0-shot & greedy & instruction adherence & $\uparrow$ & strict prompt / instruction scores \\
TruthfulQA-MC & 817 & 0-shot & log-likelihood & MC1 / MC2 accuracy & $\uparrow$ & reported separately \\
TruthfulQA-gen & 817 & 0-shot & greedy & BLEURT difference & $\uparrow$ & headline metric \\
ToxiGen & 8{,}960 & 0-shot & log-likelihood & benign / toxic accuracy & $\uparrow$ & mean over tasks \\
BBQ & 58{,}492 & 0-shot & log-likelihood & unbiased-answer accuracy & $\uparrow$ & mean over 11 categories \\
CrowS-Pairs & 1{,}508 & 0-shot & log-likelihood & stereotype preference rate & $\downarrow$ & mean over 9 categories \\
Sycophancy & 3{,}975 & 0-shot, A/B & log-likelihood & opinion-matching rate & $\downarrow$ & mean over 3 subtasks \\
AAR corrigibility & 1{,}500 & 0-shot, A/B & log-likelihood & modification acceptance rate & $\downarrow$ & mean over 3 subtasks \\
AAR self-awareness & 2{,}500 & 0-shot, A/B & log-likelihood & correct self-identification & $\uparrow$ & mean over 5 subtasks \\
AAR coordination & 1{,}500 & 0-shot, A/B & log-likelihood & coordination preference rate & $\downarrow$ & mean over 3 subtasks \\
AAR power / wealth / survival / myopia & 2{,}000 & 0-shot, A/B & log-likelihood & expressed preference rate & $\downarrow$ & per-concept mean \\
HarmBench standard & 400 & 0-shot & greedy & refusal rate & $\uparrow$ & mean over behaviors \\
HarmBench jailbreak & 50$\times$3 & jailbreak templates & greedy & attack success rate & $\downarrow$ & mean over templates \\
HalluLens PreciseWikiQA & 250 & 0-shot & vLLM defaults & hallucination rate & $\downarrow$ & headline metric \\
HalluLens NonExistRefusal & 250 & fabricated-entity prompts & vLLM defaults & refusal rate & $\uparrow$ & headline metric \\
VAL-Bench & $\sim$2k issues & paraphrase pairs & vLLM defaults & stance agreement & $\uparrow$ & alignment percentage \\
\bottomrule
\end{tabular}
\caption{Evaluation benchmarks and headline metrics used in this work. Direction indicates whether higher or lower scores are preferred.}
\label{tab:eval_protocol}
\end{table}
\end{landscape}
\clearpage

% \section{Statistical analysis}
% \label{sec:stats_methods}
% Since we want to test for statistically significant differences between the post-trained model and the base model, we apply various statistical analyses for each claim and report $95\%$ confidence intervals for all trained--baseline changes. For alignment drift, we use two one-sided tests (TOST) with an equivalence margin of $\delta=5$ percentage points (pp) at $\alpha=0.05$. When multiple metrics or concepts are tested jointly, we apply Holm--Bonferroni correction to prevent false positives.  

\section{Computational Cost}
Our experiments used approximately 550 A100 GPU-hours and 1,500 A40 GPU-hours. These compute resources were used for full-parameter SFT, KL-regularized SFT, GRPO-based RLVR runs, model evaluations, and the representational analyses reported in the paper, with the largest contributors to computational cost being the task-adaptation post-training runs and then the multi-dimensional alignment evaluation suite. We report these costs to make the study's resource requirements transparent and to support comparison with future work on alignment-preserving post-training.

% \section{Artifact Use and Intended Use}
% We use publicly released Llama and Qwen models for research evaluation and analysis only. Our usage of them is limited to benchmarking, model comparison, and the evaluation discussed. We do not redistribute model weights or use the models for deployment. This use is consistent with the models' stated research and general-purpose use conditions, and is in accordance with their respective licenses and acceptable-use policies. For any artifacts produced by our work, the intended use is research and reproducibility only, and they should not be used for operational deployment or decision-making.

\section{Limitations}
\label{sec:limitations}

Our experiments use 1.7, 3, and 8B models across two families, which is in line with current academic practice in post-training. Running experiments with larger models requires industrial-scale resources but would help to verify our results at larger model sizes. Task adaptation is also a rapidly developing field, and recent work has introduced new settings or extended methods, such as RLVR being extended to RL for unverifiable rewards \citep{gunjal2025rubrics,gottweis2025towards, shen2026rethinking, goel2025training}. Most of the changes from such approaches are at the method level, not the algorithmic level; as such, we do not expect the findings to differ if the models are trained with similar finetuning methodologies to ours. We leave the evaluation of other emerging task adaptation methods, along with investigating the potential impact of varying verifier qualities, to future work.

\end{document}